\documentclass{article}

\usepackage{epsf,conll97}

\title{A Comparative Study of the Application of Different\\ Learning Techniques to Natural Language Interfaces}
\author{Werner Winiwarter \and Yahiko Kambayashi \\
Dept.~of Information Science \\ Kyoto University \\ 
Sakyo, Kyoto 606-01, Japan \\
{\tt \{ww|yahiko\}@kuis.kyoto-u.ac.jp}}

\begin{document}
\bibliographystyle{fullname}
\maketitle

\begin{abstract}
In this paper we present first results from a comparative study. Its aim is to test the feasibility of different inductive learning techniques to perform the automatic acquisition of linguistic knowledge within a natural language database interface. In our interface architecture the machine learning module replaces an elaborate semantic analysis component. The learning module learns the correct mapping of a user's input to the corresponding database command based on a collection of past input data. We use an existing interface to a production planning and control system as evaluation and compare the results achieved by different instance-based and model-based learning algorithms.
\end{abstract}

\section{Introduction}
One of the main obstacles to the efficient use of natural language interfaces is the often required high amount of manual knowledge engineering (see \cite{Androutsopoulos+al:NLE} for a recent survey). This time-consuming and tedious process is often referred to as ``knowledge acquisition bottleneck''. It may require extensive efforts by experts highly experienced in linguistics as well as in the domain and the task \cite{Riloff+Lehnert:TOIS}. Therefore, natural language interfaces represent a domain that is very well suited for the application of machine learning algorithms to automate the acquisition process of linguistic knowledge.

So far, inductive learning has already been applied successfully to a large variety of natural language tasks. This includes basic linguistic problems such as morphological analysis \cite{Bosch+al:NEMLAP96}, parsing \cite{Zelle+Mooney:LNLP}, word sense disambiguation \cite{Mooney:EMNLP96}, and anaphora resolution \cite{Aone+Bennett:LNLP}. Besides this, there also exists some research on applications, e.g.\ machine translation \cite{Yamazaki+al:LNLP}, text categorization \cite{Moulinier+Ganascia:LNLP}, or information extraction \cite{Soderland+al:LNLP}.

The learning task in natural language interfaces is to select the correct command class based on semantic features extracted from the user input. Therefore, it can be modeled as classification problem, i.e.\ the machine learning algorithms construct a theory from the training data that is used for classifying unseen test data \cite{Quinlan:JAIR}. So far, we consider only supervised learning so that each training case has to be labeled with the correct class.

\begin{figure*}[t]
\centering \epsfile{file=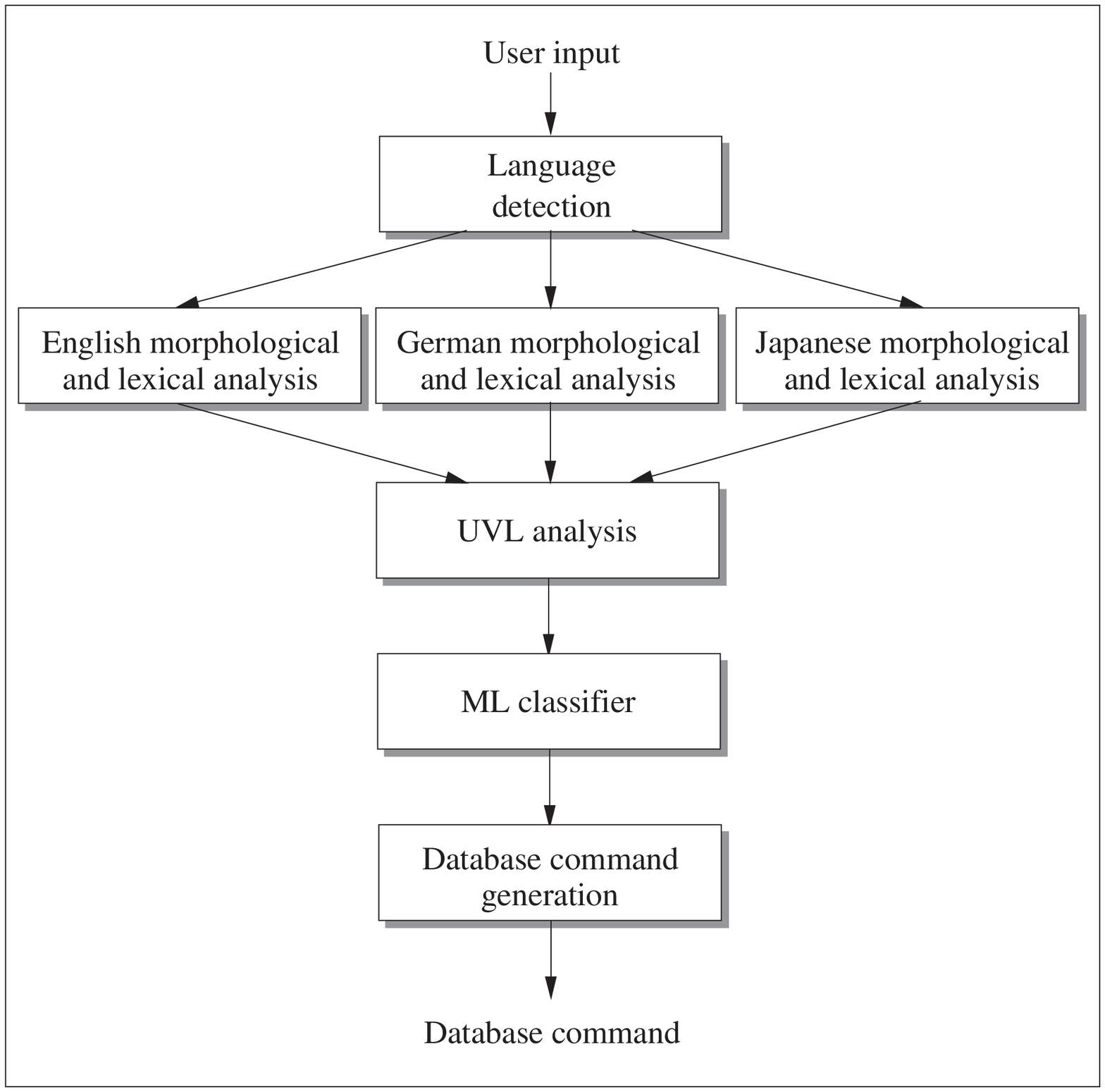,scale=0.60}
\caption{System architecture of natural language interface}                               \label{fig:1} 
\end{figure*}

We apply different existing instance-based and model-based algorithms to this problem and compare the achieved results. In addition, we have also developed several new algorithms, which we present briefly in this paper. We have implemented all algorithms by means of the deductive object-oriented database system {\em ROCK \& ROLL} \cite{Barja+al:VLDB94}. 

It solves the problem of updates in deductive databases in that it separates the declarative logic query language {\em ROLL\/} from the imperative data manipulation language {\em ROCK} within the context of a common object-oriented data model. Besides this, ROCK \& ROLL makes a clean distinction between {\em type declarations}, which describe the structural characteristics of a set of instance objects and the methods that can be applied to them, and {\em class definitions}, which specify the implementation of the methods associated with a type.

The use of the available powerful logic and object-oriented programming language enables an efficient implementation of the different approaches to machine learning. It also gives us a convenient integrated tool that assists in applying the machine learning algorithms to the data collection stored in the same database.

As comparative evaluation of the implemented algorithms, we applied them to an extensive case study: a natural language interface for a production planning and control system. The system is used in a multilingual environment, which includes the languages English, German, and Japanese. Therefore, an important issue of the evaluation was to check whether the learned knowledge is language-independent, i.e.\ if it really operates based on semantic deep forms so that it abstracts from linguistic surface phenomena.

The rest of the paper is organized as follows. First, we briefly introduce the learning task before we present the applied machine learning algorithms in more detail. Finally, we explain the set-up of the case study and discuss the achieved results from evaluation.

\section{Learning Task}
\label{lt}
Our interface architecture is displayed in Fig.~\ref{fig:1}. It represents a multilingual database interface for the languages English, German, and Japanese. First, the {\em language} of the user input is {\em detected} and the input is transferred to the corresponding language-specific {\em morphological and lexical analyzer}.

Morphological and lexical analysis performs the {\em tokenization} of the input, i.e.\ the segmentation into individual words or tokens. This task is not always trivial as in the case of Japanese, which uses no spaces for separating words. As next step the input is transformed into a {\em deep form list (DFL)}, which indicates for each token its surface form, category, and semantic deep form. 

\begin{figure*}[t]
\centering \epsfile{file=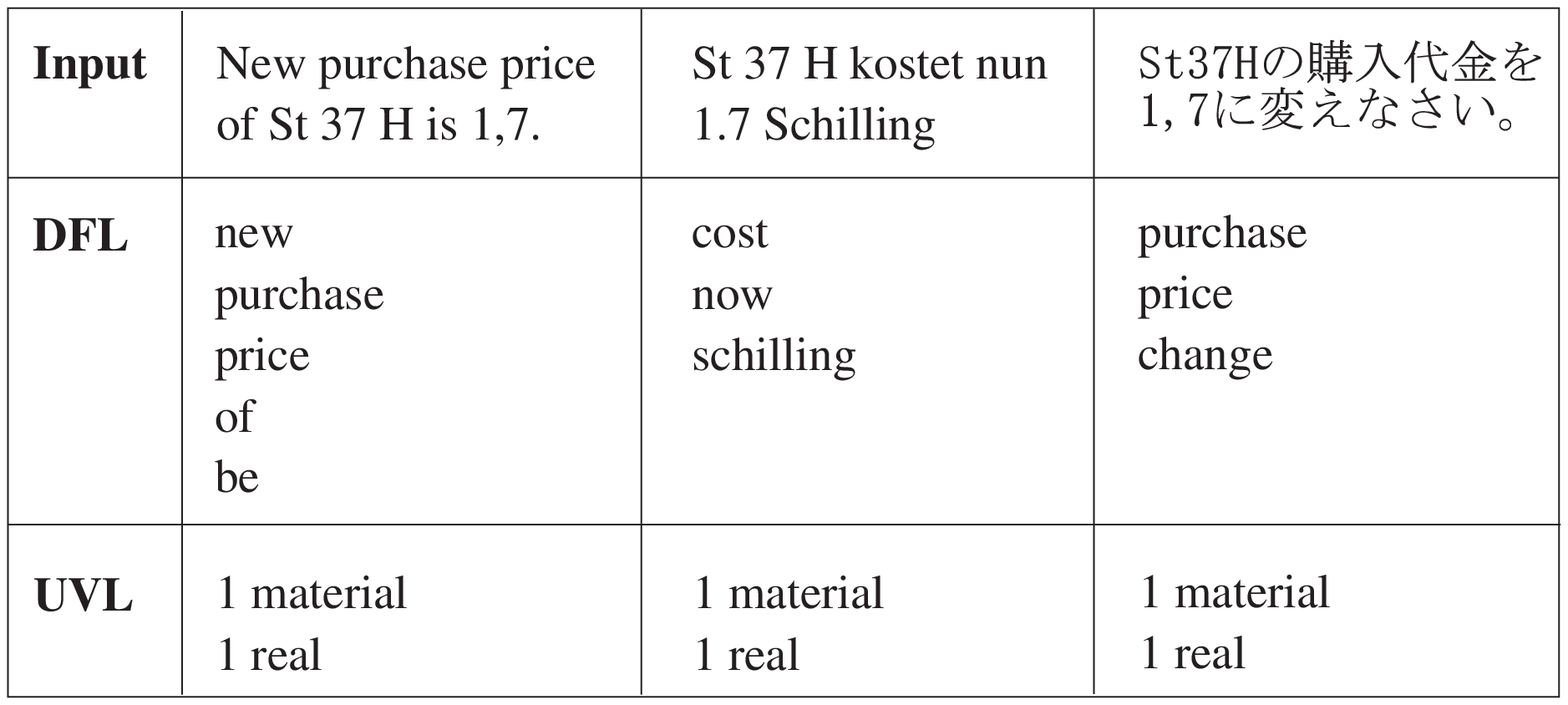,scale=0.60}
\caption{Example of feature encoding}                               \label{fig:2} 
\end{figure*}

For database interfaces, unknown values contained in the input possess particular importance for the meaning of a command. Therefore, we treat those unknown values separately in the {\em unknown value list (UVL) analyzer}. This module checks the data type of unknown values and looks them up in the database to find out whether they represent identifiers of existing entities. In such a case, the entity type is indicated in the resulting UVL, otherwise we use the data type instead.

DFL and UVL represent the input to the {\em machine learning (ML) classifier}. It assigns a ranked list of command classes to the input sentence according to the learned classification rules. As last step the classifications are used for {\em generating} appropriate {\em database commands}.

For the encoding of the training data we only make use of the semantic deep forms contained in the DFL. We use English concepts as deep forms and map them to binary features, i.e.\ a certain feature equals $1$ if the deep form is a member of the DFL, otherwise it equals $0$. For the elements of the UVL we apply a more detailed encoding, which maps the number and the type to binary features. Figure~\ref{fig:2} shows an example of the features derived from English, German, and Japanese input sentences for the update of the purchase price for a material.

Thus, the learning task replaces an elaborate semantic analysis of the user input. The development of the corresponding underlying rule base might require several man-months. The learning task represents a realistic real-life application, which differs from many other problems studied in machine learning research in that it consists of a large number of features and classes. Furthermore, the command classes are often very similar and even for human experts very difficult to distinguish.

\section{Learning Algorithms}
\subsection{Instance-based Learning}
Instance-based approaches represent the learned knowledge simply as collection of training cases or {\em instances}. For that purpose they use the same language as for the description of the training data \cite{Quinlan:ML93}. A new case is then classified by finding the instance with the highest similarity and using its class as prediction. Therefore, instance-based algorithms are characterized by a very low training effort. On the other hand, this leads to a high storage requirement because the algorithm has to keep all training cases in memory. Besides this, one has to compare new cases with all existing instances, which results in a high computation cost for classification.

Different instance-based algorithms vary in how they assess the similarity (or distance) between two instances. Two very commonly used methods are {\em IB1} \cite{Aha+al:ML} and {\em IB1-IG} \cite{Daelemans+Bosch:TWLT3}. Whereas IB1 applies the simple approach of treating all features as equally important, IB1-IG uses the information gain \cite{Quinlan:ML} of the features as weighting function.

We have developed an algorithm called {\em BIN-CAT} for binary features with class-dependent weighting and asymmetric treatment of the feature values. The similarity between a new case $X$ and a training case~$Y$ is calculated according to the following formula:

\begin{eqnarray}
{\rm SIM}_{X,Y} & = & \sum_{i=1}^{n} p \left( D_{i},C_{Y} \right) \cdot w_{i} \cdot \sigma \left( x_{i},y_{i} \right) - \nonumber \\
& & \sum_{i=1}^{n} p \left( D_{i},C_{Y} \right) \cdot w_{i} \cdot \delta_{Y} \left( x_{i},y_{i} \right) - \nonumber \\
& & \sum_{i=1}^{n} \left[ 1 - p \left( D_{i},C_{Y} \right) \right] \cdot w_{i} \cdot \delta_{X} \left( x_{i},y_{i} \right) \enspace . \nonumber \\
 & & 
\label{eq:1}
\end{eqnarray}

In this formula, $n$ indicates the number of features, $D_{i}$ the number of instances that have value $1$ for feature $i$, and $C_{Y}$ the class of the training case~$Y$. The term $p(D_{i},C_{Y})$ then denotes the proportion of instances in $D_{i}$ that belong to class $C_{Y}$. $\sigma (x_{i},y_{i})$, $\delta_{Y}(x_{i},y_{i})$, and $\delta_{X}(x_{i},y_{i})$ are determined as follows: 

\begin{eqnarray}
\sigma \left( x_{i},y_{i} \right) & = & \left\{
\begin{array}{ll}
1 & \mbox{if } x_{i}=1 \wedge y_{i}=1 \\
0 & {\rm otherwise}
\end{array} \right. \nonumber \\
\delta_{Y} \left( x_{i},y_{i} \right) & = & \left\{
\begin{array}{ll}
1 & \mbox{if } x_{i}=0 \wedge y_{i}=1 \\
0 & {\rm otherwise}
\end{array} \right. \nonumber \\
\delta_{X} \left( x_{i},y_{i} \right) & = & \left\{
\begin{array}{ll}
1 & \mbox{if } x_{i}=1 \wedge y_{i}=0 \\
0 & {\rm otherwise}
\end{array} \right. 
\label{eq:2}
\end{eqnarray}
so that the second sum in (\ref{eq:1}) is rated higher for a larger number of occurrences of the $i$th feature for class $C_{Y}$ whereas the third sum is rated lower. This means that if the training case $Y$ contains a certain feature and the new case $X$ does not, then we rate this difference the stronger the more often the feature occurs for class $C_{Y}$. On the other hand, for features appearing in the new case $X$ but not in~$Y$, the opposite is true.

Finally, $w_{i}$ represents the weight of feature $i$. It is calculated by making use of the following formula:

\begin{equation}
w_{i}=\frac{1}{c} \cdot \sum_{j=1}^{c} 1 - 4 \cdot p(D_{i},j) \cdot \left[ 1 - p(D_{i},j) \right] \enspace .
\label{eq:3}
\end{equation}

The term under the summation symbol represents the selectivity of feature $i$ for class $j$. It equals $1$ if either all or none of the cases have value $1$ for this feature. In other words, all instances for class~$j$ then either possess or do not possess this feature, which makes it a very discriminative characteristic. The other extreme is that $p(D_{i},j)$ equals 50\,\%. In that case, this feature allows for no prediction of the class and the term under the summation symbol becomes~$0$.

\begin{figure*}[t]
\centering \epsfile{file=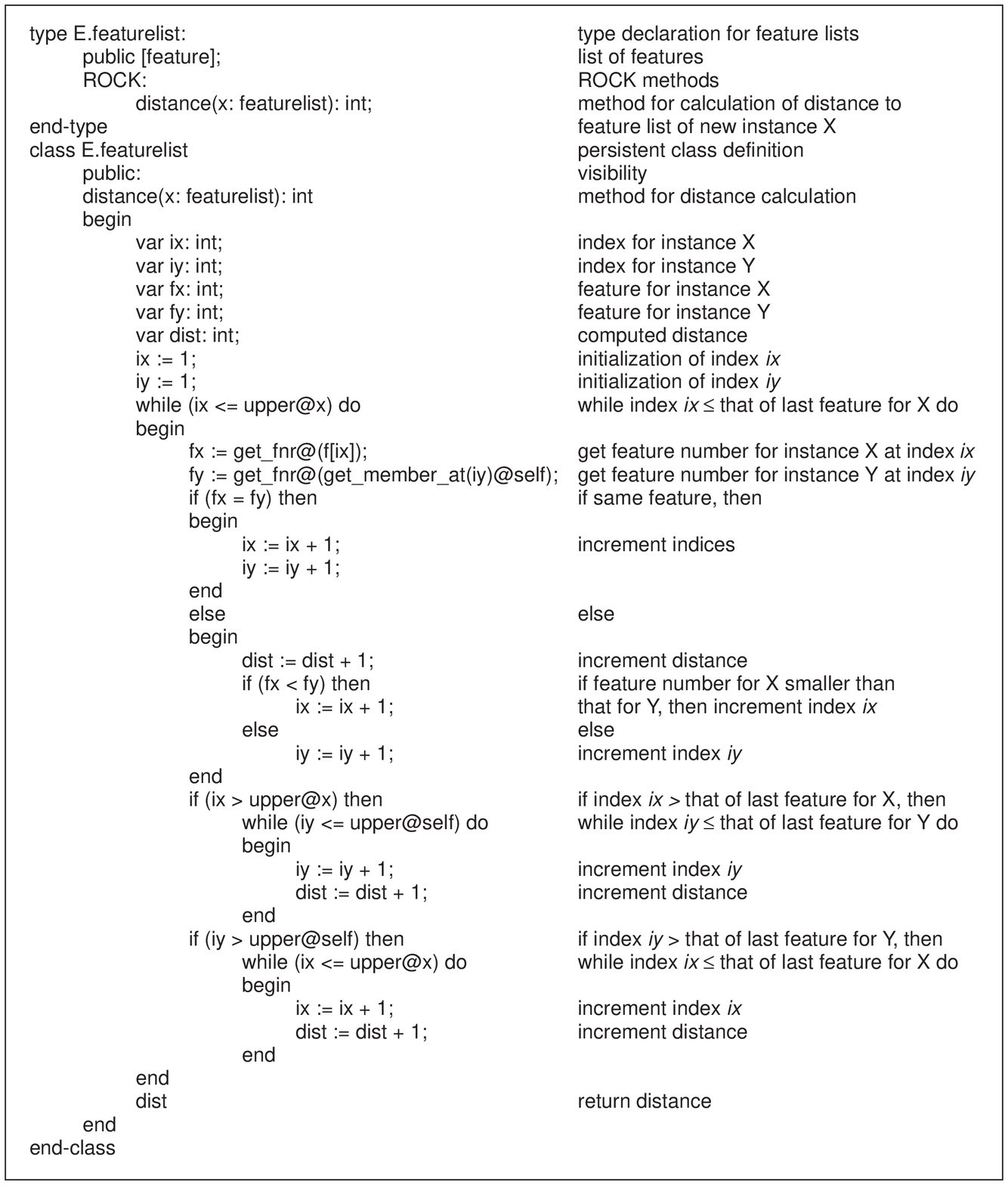,scale=0.68}
\caption{ROCK \& ROLL code segment for IB1 distance calculation}
\label{fig:3} 
\end{figure*}

We have implemented all above-mentioned algorithms for binary features in ROCK \& ROLL in that we store the instances as objects and assign to them the features as ordered lists sorted by the feature numbers. The calculation of the similarity between two cases is then realized as method invocation on the feature list. For example, Fig.~\ref{fig:3} shows the ROCK method to compute the distance between two feature lists according to IB1.

Besides pure instance-based learning we have also developed an algorithm {\em BIN-PRO}, which creates a {\em prototype} for each class. Those prototypes are then used for the comparison with new cases. This has the big advantage that one does not have to store all the training instances and that the number of required comparisons for classification is reduced to the number of existing classes. As similarity function between a new case $X$ and a certain class $C$ we use the following formula:

\begin{eqnarray}
{\rm SIM}_{X,C} & = & \sum_{f \in X} \left| D_{C} \right| \cdot p \left( D_{f},C \right) \cdot w_{f} - \nonumber \\
& & \sum_{f \not \in X} p \left( D_{f},C \right) \cdot w_{f} \enspace . \label{eq:4}
\end{eqnarray}

In this formula, we give more emphasis to features~$f$ that are present in $X$ in that we multiply them by $\left| D_{C} \right|$, the number of instances for class $C$. However, the second sum takes also important features for class $C$ into account that are missing in the new case $X$. As weighting function $w_{f}$ we use again~(\ref{eq:3}). The implementation in ROCK \& ROLL is performed by creating an object for each prototype and by invoking the associated method for computing the similarity to a new test case.

\subsection{Model-based Learning}
\label{mbl}
In contrast to instance-based learning, model-based approaches represent the learned knowledge in a theory language that is richer than the language used for the description of the training data \cite{Quinlan:ML}. Such learning methods construct explicit generalizations of training cases resulting in a large reduction of the size of the stored knowledge base and the cost of testing new test cases.

In our research we consider the subtypes of decision trees and rule-based learning as well as hybrid approaches between them. The main difference between the various methods for constructing {\em decision trees} is the selection of the feature for splitting a node. The following two main categories are distinguished:

\begin{itemize}
\item {\em static splitting}: selects the best feature for splitting always on the basis of the complete collection of instances,
\item {\em dynamic splitting}: re-evaluates the best feature for splitting for each node based on the current local set of instances.
\end{itemize}

Static splitting requires less computational effort because it performs the feature ranking only once for the construction process. However, it entails overhead to keep track of already used features and to eliminate features that provide no proper splitting of the set of instances. Besides that, dynamic splitting methods produce much more compact trees with fewer nodes, leaves, and levels. This results in a sharp reduction of the storage requirement as well as the number of comparisons during classification.

We have implemented decision trees for static ({\em BS-tree}) and dynamic splitting ({\em BD-tree}) by using the weighting function (\ref{eq:3}) as ranking scheme for the splitting criterion. In addition, we have also implemented the {\em IGTree} algorithm \cite{Daelemans+al:AIR}, which uses the information gain as static splitting criterion, and {\em C4.5} \cite{Quinlan:C4.5}, which applies the information gain to dynamic splitting. The decision trees are implemented in ROCK~\&~ROLL by creating an object for each node and by linking the nodes according to the tree structure. The classification of a new case is then simply performed as top-down traversal of the tree starting from the root. Besides this exact search we have also implemented an approximate search method, which allows one incorrect edge along the traversal to find a larger number of similar cases.

{\em Rule-based learning} represents a second large category of model-based techniques. It aims at deriving a set of rules from the instances of the training set. A {\em rule} is here defined as a conjunction of {\em literals}, which, if satisfied, assigns a class to a new case. For the case of binary features, the literals correspond to {\em feature tests} with positive or negative {\em sign}. This means that they check whether a new case possesses a certain feature (for positive tests) or not (for negative tests).

\begin{figure*}[t]
\centering \epsfile{file=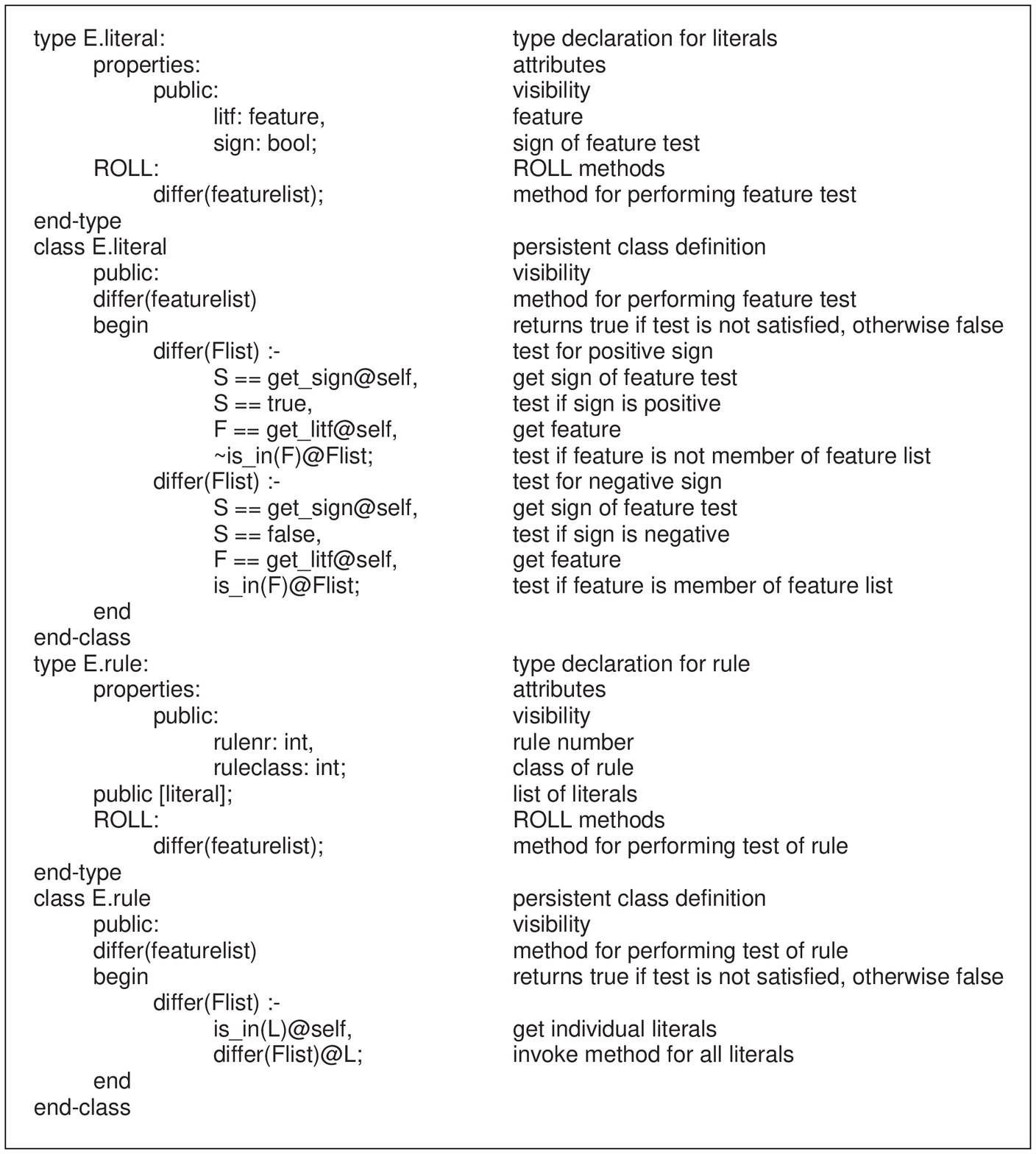,scale=0.68}
\caption{ROCK \& ROLL code segment for test of rules}
\label{fig:4} 
\end{figure*}

The methods for deriving the rules originate from the field of {\em inductive logic programming} \cite{Muggleton:ILP}. One of the most prominent algorithms for rule-based learning is {\em FOIL} \cite{Quinlan+Cameron-Jones:NGC}, which learns for each class a set of rules by applying a separate-and-conquer strategy. The algorithm takes the instances of a certain class as {\em target relation}. It iteratively learns a rule and removes those instances from the target relation that are covered by the rule. This is repeated until no instances are left in the target relation. A rule is grown by repeated specialization, adding literals until the rule does not cover any instances of other classes. In other words, the algorithm tries to find rules that possess some {\em positive bindings}, i.e.\ instances that belong to the target relation, but no {\em negative bindings} for instances of other classes. Therefore, the reason for adding a literal is to increase the relative proportion of positive bindings.

As weighting function for selecting the next literal, FOIL uses the information gain. We have implemented FOIL, and besides this, we also use the algorithm {\em BIN-rules} with the following weighting function:

\begin{equation}
W_{f,s,C}=b_{f}^{+} \cdot (b^{-} - b_{f}^{-}) \cdot w_{f,s,C} \enspace .
\label{eq:5}
\end{equation}

In this formula, $s$ indicates the sign of the feature test. The number of positive (negative) bindings after adding the literal for the test of feature $f$ is written as $b_{f}^{+}$ ($b_{f}^{-}$). Finally, $b^{-}$ indicates the number of negative bindings before adding the literal so that $b^{-} - b_{f}^{-}$ calculates the reduction of negative bindings achieved by adding the literal. The weights $w_{f,s,C}$ are calculated as class-dependent weights for class $C$ by making use of the feature weights $w_{f}$ from (\ref{eq:3}):

\begin{equation}
w_{f,s,C} = \left\{
\begin{array}{ll}
w_{f} \cdot p(D_{f},C) & \mbox{if $s$ positive} \\
w_{f} \cdot \left[ 1 - p(D_{f},C) \right] & {\rm otherwise} \enspace . 
\end{array}
\right.
\label{eq:6}
\end{equation}

We have implemented the test of rules as deductive ROLL method as shown in Fig.~\ref{fig:4}. The invocation of the method is a query with the parameter {\tt fl} for the feature list of the new case. The test returns {\tt false} for those rules that are satisfied by the new case. The result of the query can then be assigned to the set of satisfied rules {\tt rs} by using the command: {\tt rs~:= [\{R\}|$\sim$differ(!fl)@R];}. As in the case of decision trees, we have developed an approximate test, which tolerates one divergent literal.

As last group of model-based algorithms we look at {\em hybrid approaches} between decision trees and rule-based learning. There exist two ways in principle to combine the advantages of the two paradigms. The first one is to extract rules from a decision tree whereas the second one follows the opposite direction by constructing a decision tree from a rule base. 

As example of the first type of approach we have implemented {\em C4.5-RULES} \cite{Quinlan:C4.5}, which extracts rules from the decision tree built by C4.5. Rules are computed as paths along the traversal from the root to all leaves. In a second run, rules are pruned by removing redundant literals and rules.

Regarding the second type of approach, we start from the rule base produced by BIN-rules and use it for building an {\em SE-tree} \cite{Rymon:ML93}. SE-trees are a generalization of decision trees in that they allow not only one but several feature tests at one node. Therefore, a much flatter and more compact tree structure is achieved. For the construction of the tree we sort the feature tests of the rules first. Starting from a root node, we then construct paths according to the literals of the individual rules. For this process we make use of existing paths as far as possible before creating new branches.

\section{Evaluation}
As case study for investigating the feasibility of the implemented machine learning algorithms, we use a multilingual natural language interface to a {\em production planning and control system (PPC)}. The PPC performs the mean-term scheduling of products and resources involved in the manufacturing processes, i.e.\ material, machines, and labor. The resulting master production schedule forms the basis of the coordination of related business services such as engineering, manufacturing, and finance. The modeled enterprise makes precision tools by using job order production and serial manufacture as basic strategies. The efficient realization of the high demands of the application exceeds the power of relational database technology. Therefore, it represents an excellent choice for deriving full advantage of the extended functionality of deductive object-oriented database systems. Furthermore, the sophisticated functionality justifies the effective use of a natural language interface.

During previous research \cite{Winiwarter:IDA} we developed a German natural language interface based on 1000 input sentences that had been collected from users by means of questionnaires. The input sentences were then mapped to 100 command classes (10 for each class). The mapping was performed by elaborate semantic analysis; for the development of the underlying rule base we spent several man-months.

Therefore, we were eager to see if we could replace this extensive effort by a machine learning component that learns the same linguistic knowledge automatically. For this purpose we divided the 1000 sentences into 900 training cases and 100 test cases. In addition, we collected 100 Japanese and 100 English test sentences to check whether the learned knowledge really operates at a semantic level independent from language-specific phenomena.

As result of the encoding of the training set (see Sect.~\ref{lt}), we obtained the large number of 316 features, 289 for the DFL and 27 for the UVL. For the evaluation of the different machine learning algorithms we used as performance measures the {\em success rate}, i.e.\ the proportion of correctly classified test cases, and the {\em top-3 rate}. The latter indicates the proportion of cases where the correct classification is among the first three predicted classes. For the case of model-based approaches we had to produce additional candidates for classes. This was achieved by applying approximate methods that allow one incorrect edge along the traversal of decision trees or one divergent literal for the test of rules (see Sect.~\ref{mbl}).

\begin{table*}[t] 
\begin{center}       
\begin{tabular}{|c|c|c|c|c|c|c|} \hline
 & \multicolumn{2}{|c|}{GERMAN} & 
\multicolumn{2}{|c|}{ENGLISH} & 
\multicolumn{2}{|c|}{JAPANESE}\\ \hline
  & Success rate & Top-3 rate & Success rate & Top-3 rate & Success rate & Top-3 rate \\ \hline 
IB1     & 82\,\% &  94\,\% & 98\,\% &  99\,\% & 94\,\% &  98\,\% \\ \hline
IB1-IG  & 84\,\% &  98\,\% & 97\,\% & 100\,\% & 90\,\% &  99\,\% \\ \hline
BIN-CAT & 94\,\% & 100\,\% & 99\,\% & 100\,\% & 99\,\% & 100\,\% \\ \hline
BIN-PRO & 95\,\% & 100\,\% & 97\,\% & 100\,\% & 97\,\% & 100\,\% \\ 
\hline
\end{tabular}
\caption{Test results for instance-based learning} 
\label{tab:1} 
\end{center}             
\end{table*}

\begin{table*}[t] 
\begin{center}       
\begin{tabular}{|c|c|c|c|c|c|c|} \hline
 & \multicolumn{2}{|c|}{GERMAN} & 
\multicolumn{2}{|c|}{ENGLISH} & 
\multicolumn{2}{|c|}{JAPANESE}\\ \hline
  & Success rate & Top-3 rate & Success rate & Top-3 rate & Success rate & Top-3 rate \\ \hline 
IGTree  & 80\,\% &  94\,\% & 92\,\% & 100\,\% & 86\,\% &  97\,\% \\ \hline
BS-tree & 86\,\% &  97\,\% & 95\,\% & 100\,\% & 90\,\% &  96\,\% \\ \hline 
C4.5    & 94\,\% & 100\,\% & 94\,\% & 100\,\% & 89\,\% & 100\,\% \\ \hline
BD-tree & 93\,\% &  99\,\% & 94\,\% &  99\,\% & 91\,\% &  99\,\% \\ \hline
SE-tree & 94\,\% &  97\,\% & 96\,\% &  97\,\% & 91\,\% &  95\,\% \\ \hline
\end{tabular}
\caption{Test results for decision trees} 
\label{tab:2}        
\end{center}      
\end{table*}

Our first experiment was the comparison of the four instance-based algorithms IB1, IB1-IG, BIN-CAT, and BIN-PRO. As can be seen from the results in Table~\ref{tab:1}, BIN-CAT clearly outperforms IB1 and IB1-IG. Concerning the method BIN-PRO, which uses prototypes of classes, we achieved results at the same quality level as for BIN-CAT. This is remarkable if one considers the much more condensed representation of the learned knowledge.

The comparison between the results for the individual languages shows that there is no advantage for the German test sentences. On the contrary, the test results for German are inferior to that for English or Japanese. This may be partly due to a greater deviation of the German expressions and phrases used in the test set from the ones used in the training set. Besides this, the restriction of extracted features during encoding the test set to those learned from the training set certainly performs an important filtering function. It removes language-specific syntactic particles that do not contribute to the meaning of the input. This is especially true for the case of Japanese sentences, which possess a completely different syntactic structure in comparison with English or German including many particles with no equivalent words in the other two languages. 

The second part of the evaluation was the comparison of the four algorithms for building decision trees: IGTree, BS-tree, C4.5, and BD-tree. Besides this, we also included the SE-tree constructed by a hybrid approach (see Sect.~\ref{mbl}). The test results in Table~\ref{tab:2} indicate that the trees with dynamic splitting are superior to those with static splitting and that C4.5, BD-tree, and SE-tree produce results of similar quality. Table~\ref{tab:3} compares the number of nodes, leaves, and levels for the individual trees. The two trees with dynamic splitting are much more compact than those with static splitting, with C4.5 clearly outperforming BD-tree. Finally, the hybrid SE-tree is much flatter than C4.5 but possesses a larger number of nodes and leaves. 

\begin{table}[ht] 
\begin{center}       
\begin{tabular}{|c|c|c|c|} \hline
  & Nodes & Leaves & Levels \\ \hline 
IGTree  & 865 & 433 & 33 \\ \hline 
BS-tree & 719 & 360 & 86 \\ \hline
C4.5    & 339 & 170 & 26 \\ \hline
BD-tree & 451 & 226 & 52 \\ \hline 
SE-tree & 559 & 209 &  8 \\ \hline
\end{tabular}
\caption{Characteristics for decision trees} 
\label{tab:3}              
\end{center}
\end{table}

As last part of our comparative study we tested the rule-based techniques FOIL, BIN-rules, and the hybrid approach C4.5-RULES. As Table~\ref{tab:4} shows, FOIL produces the most compact representation of learned knowledge, followed by C4.5-RULES and BIN-rules. However, according to Table~\ref{tab:5} both BIN-rules and C4.5-RULES outperform FOIL with almost identical results.

\begin{table}[ht] 
\begin{center}       
\begin{tabular}{|c|c|c|c|} \hline
  & Rules & Literals & Max. length \\ \hline 
FOIL       & 215 & 534 &  5 \\ \hline
BIN-rules  & 209 & 726 &  7 \\ \hline 
C4.5-RULES & 167 & 677 & 24 \\ \hline
\end{tabular}
\caption{Characteristics for rule-based learning} 
\label{tab:4}        
\end{center}      
\end{table}

\begin{table*}[t] 
\begin{center}       
\begin{tabular}{|c|c|c|c|c|c|c|} \hline
 & \multicolumn{2}{|c|}{GERMAN} & 
\multicolumn{2}{|c|}{ENGLISH} & 
\multicolumn{2}{|c|}{JAPANESE}\\ \hline
  & Success rate & Top-3 rate & Success rate & Top-3 rate & Success rate & Top-3 rate \\ \hline 
FOIL       & 85\,\% & 97\,\% & 92\,\% & 97\,\% & 88\,\% & 96\,\% \\ \hline
BIN-rules  & 94\,\% & 97\,\% & 95\,\% & 97\,\% & 91\,\% & 95\,\% \\ \hline
C4.5-RULES & 94\,\% & 98\,\% & 94\,\% & 96\,\% & 91\,\% & 96\,\% \\ \hline
\end{tabular}
\caption{Test results for rule-based learning} 
\label{tab:5}
\end{center}
\end{table*}

\begin{figure*}[t]
\centering \epsfile{file=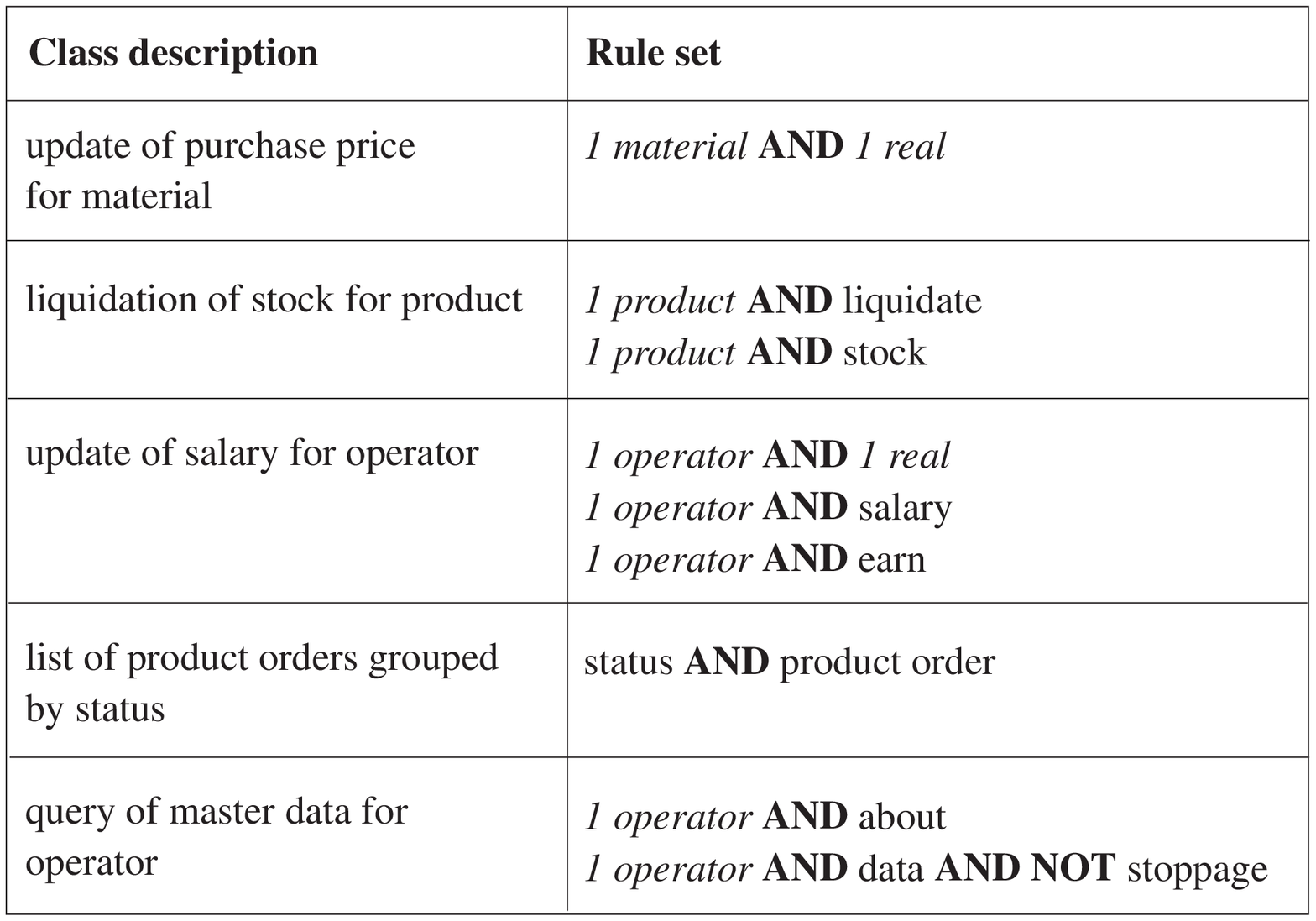,scale=0.60}
\caption{Examples of learned rules}
\label{fig:5} 
\end{figure*}

An advantage of rule-based learning in comparison with other methods is that the learned knowledge can be easily presented to the user in a clear and understandable form. The derived rules allow a transparent knowledge representation that one can use for explaining decisions of the system to the user. Figure~\ref{fig:5} gives some examples of rule sets learned by BIN-rules for several command classes.

If we take a final look at Table~\ref{tab:1}, Table~\ref{tab:2}, and Table~\ref{tab:5}, we can see that independent from the applied machine learning paradigm the achieved results reached satisfactory quality for all three groups. By considering the three best representatives BIN-CAT, C4.5, and BIN-rules, we obtain an average success rate for all three languages of 94.3\,\% and a top-3 rate of 98.8\,\%. This result is surprisingly high if one considers the complexity of the task at hand. Unfortunately, we had no possibility of a direct comparison with the results of the hand-engineered interface because the previous interface had been developed only for German based on the complete collection of 1000 sentences by using a different software. In any case, we could show that machine learning represents a sound alternative to manual knowledge acquisition for the application in natural language interfaces.

\section{Conclusion}
In this paper we have presented first results from a comparative study of applying different inductive learning techniques to natural language interfaces. We have implemented a representative selection of instance-based and model-based algorithms by making use of deductive object-oriented database functionality. The extensive case study for an interface to a production planning and control system shows the feasibility of the approach in that linguistic knowledge is learned the acquisition of which normally takes a large effort of human experts.

Future work will concentrate on the important point of increasing the reliability of test results in that we apply cross-validation trials and statistical tests for the significance of performance differences between two algorithms. Furthermore, we also want to generate learning functions that plot success rates as function of the size of the training collection. Besides this, we plan to test our learning algorithms on standard benchmark machine learning datasets and other typical natural language learning datasets.

Finally, we intend to extend the implemented algorithms to include also unsupervised methods as well as connectionist and evolutionary techniques. In addition, we will implement incremental learning techniques, which continue the learning process during the test phase, and adaptive boosting methods, which apply several classifiers instead of just one. We believe that our study is a first promising step towards the challenging task of carrying out comparative evaluations of the performance of different machine learning algorithms for specific linguistic problems.


\begin{thebibliography}{fullname}

\bibitem[\protect\citename{Aha \bgroup et al.\egroup }1991]{Aha+al:ML}
David~W. Aha, Dennis Kibler, and Marc Albert.
\newblock 1991.
\newblock Instance-based learning algorithms.
\newblock {\em Machine Learning}, 7:37--66.

\bibitem[\protect\citename{Androutsopoulos \bgroup et al.\egroup }1995]{Androutsopoulos+al:NLE}
Ioannis Androutsopoulos, Graeme~D. Ritchie, and Peter Thanisch.
\newblock 1995.
\newblock Natural language interfaces to databases --- an introduction.
\newblock {\em Journal of Natural Language Engineering}, 1(1):29--81.

\bibitem[\protect\citename{Aone and Bennett}1996]{Aone+Bennett:LNLP}
Chinatsu Aone and Scott~W. Bennett.
\newblock 1996.
\newblock Applying machine learning to anaphora resolution.
\newblock In S.~Wermter, E.~Riloff, and G.~Scheler, editors, {\em 
Connectionist, Statistical, and Symbolic Approaches to Learning for Natural Language Processing}, pages 302--314. Springer-Verlag, Berlin, Germany.

\bibitem[\protect\citename{Barja \bgroup et al.\egroup }1994]{Barja+al:VLDB94}
Maria~L. Barja, Norman~W. Paton, Alvaro~A.A. Fernandes, M.~Howard Williams, and Andrew Dinn.
\newblock 1994.
\newblock An effective deductive object-oriented database through language integration.
\newblock In {\em Proceedings of the 20th International Conference on Very Large Data Bases}, pages 463--474, Athens, Greece. Morgan Kaufmann, San Mateo, California.

\bibitem[\protect\citename{Daelemans and van den Bosch}1992]{Daelemans+Bosch:TWLT3}
Walter Daelemans and Antal van den Bosch.
\newblock 1992.
\newblock Generalisation performance of backpropagation learning on a syllabification task.
\newblock In M.~Drossaers and A.~Nijholt, editors, {\em 
TWLT3: Connectionism and Natural Language Processing}, pages 27--37. Twente University Press, Enschede, Netherlands.

\bibitem[\protect\citename{Daelemans \bgroup et al.\egroup }1997]{Daelemans+al:AIR}
Walter Daelemans, Antal van den Bosch, and Ton Weijters.
\newblock 1997.
\newblock {IGT}ree: Using trees for compression and classification in lazy learning algorithms.
\newblock {\em Artificial Intelligence Review}.
\newblock To appear.

\bibitem[\protect\citename{Mooney}1996]{Mooney:EMNLP96}
Raymond~J. Mooney.
\newblock 1996.
\newblock Comparative experiments on disambiguating word senses: An illustration of the role of bias in machine learning.
\newblock In {\em Proceedings of the Conference on Empirical Methods in Natural Language Processing}, pages 82--91, Philadelphia, Pennsylvania, May.

\bibitem[\protect\citename{Moulinier and Ganascia}1996]{Moulinier+Ganascia:LNLP}
Isabelle Moulinier and Jean-Gabriel Ganascia.
\newblock 1996.
\newblock Applying an existing machine learning algorithm to text categorization.
\newblock In S.~Wermter, E.~Riloff, and G.~Scheler, editors, {\em 
Connectionist, Statistical, and Symbolic Approaches to Learning for Natural Language Processing}, pages 343--354. Springer-Verlag, Berlin, Germany.

\bibitem[\protect\citename{Muggleton}1992]{Muggleton:ILP}
Stephen Muggleton, editor.
\newblock 1992.
\newblock {\em Inductive Logic Programming}. Academic Press, London, England. 

\bibitem[\protect\citename{Quinlan}1986]{Quinlan:ML}
J.~Ross Quinlan.
\newblock 1986.
\newblock Induction of decision trees.
\newblock {\em Machine Learning}, 1:81--206.

\bibitem[\protect\citename{Quinlan}1993a]{Quinlan:ML93}
J.~Ross Quinlan.
\newblock 1993a.
\newblock Combining instance-based and model-based learning.
\newblock In {\em Proceedings of the 10th International Conference on Machine Learning}, pages 236--243, Amherst, Massachusetts. Morgan Kaufmann, San Mateo, California. 

\bibitem[\protect\citename{Quinlan}1993b]{Quinlan:C4.5}
J.~Ross Quinlan.
\newblock 1993b.
\newblock {\em C4.5: Programs for Machine Learning}. Morgan Kaufmann, San Mateo, California.

\bibitem[\protect\citename{Quinlan and Cameron-Jones}1995]{Quinlan+Cameron-Jones:NGC}
J.~Ross Quinlan and R.~Michael Cameron-Jones.
\newblock 1995.
\newblock Induction of logic programs: {FOIL} and related systems.
\newblock {\em New Generation Computing}, 13:287--312.

\bibitem[\protect\citename{Quinlan}1996]{Quinlan:JAIR}
J.~Ross Quinlan.
\newblock 1996.
\newblock Learning first-order definitions of functions.
\newblock {\em Journal of Artificial Intelligence Research}, 5:139--161.

\bibitem[\protect\citename{Riloff and Lehnert}1994]{Riloff+Lehnert:TOIS}
Ellen Riloff and Wendy Lehnert.
\newblock 1994.
\newblock Information extraction as a basis for high-precision text classification.
\newblock {\em ACM Transactions on Information Systems}, 12(3):296--333.

\bibitem[\protect\citename{Rymon}1993]{Rymon:ML93}
Ron Rymon.
\newblock 1993.
\newblock An {SE}-tree-based characterization of the induction problem.
\newblock In {\em Proceedings of the 10th International Conference on Machine Learning}, pages 268--275, Amherst, Massachusetts. Morgan Kaufmann, San Mateo, California. 

\bibitem[\protect\citename{Soderland \bgroup et al.\egroup }1996]{Soderland+al:LNLP}
Stephen Soderland, David Fisher, Jonathan Aseltine, and Wendy Lehnert.
\newblock 1996.
\newblock Issues in inductive learning of domain-specific text extraction rules.
\newblock In S.~Wermter, E.~Riloff, and G.~Scheler, editors, {\em 
Connectionist, Statistical, and Symbolic Approaches to Learning for Natural Language Processing}, pages 290--301. Springer-Verlag, Berlin, Germany.

\bibitem[\protect\citename{van den Bosch \bgroup et al.\egroup }1996]{Bosch+al:NEMLAP96}
Antal van den Bosch, Walter Daelemans, and Ton Weijters.
\newblock 1996.
\newblock Morphological analysis as classification: An inductive-learning approach.
\newblock In {\em Proceedings of the Second International Conference on New Methods in Language Processing}, Ankara, Turkey, September.

\bibitem[\protect\citename{Winiwarter}1994]{Winiwarter:IDA}
Werner Winiwarter.
\newblock 1994.
\newblock {\em The Integrated Deductive Approach to Natural Language Interfaces}. PhD thesis, University of Vienna, Austria.

\bibitem[\protect\citename{Yamazaki \bgroup et al.\egroup }1996]{Yamazaki+al:LNLP}
Takefumi Yamazaki, Michael~J. Pazzani, and Christopher Merz.
\newblock 1996.
\newblock Acquiring and updating hierarchical knowledge for machine translation based on a clustering technique.
\newblock In S.~Wermter, E.~Riloff, and G.~Scheler, editors, {\em 
Connectionist, Statistical, and Symbolic Approaches to Learning for Natural Language Processing}, pages 329--342. Springer-Verlag, Berlin, Germany.

\bibitem[\protect\citename{Zelle and Mooney}1996]{Zelle+Mooney:LNLP}
John M. Zelle and Raymond J. Mooney.
\newblock 1996.
\newblock Comparative results on using inductive logic programming for corpus-based parser construction.
\newblock In S.~Wermter, E.~Riloff, and G.~Scheler, editors, {\em 
Connectionist, Statistical, and Symbolic Approaches to Learning for Natural Language Processing}, pages 355--369. Springer-Verlag, Berlin, Germany.

\end{thebibliography}
\end{document}